\documentclass[letterpaper, 10 pt, conference]{ieeeconf}  
\usepackage{url}
\usepackage{cite}
\usepackage{graphicx} 
\usepackage{epsfig} 
\usepackage{mathptmx} 
\usepackage{times} 
\usepackage{amsmath} 
\usepackage{amssymb}  
\usepackage{gensymb}
\usepackage{epstopdf}
\usepackage[]{units}
\usepackage{color}
\usepackage{verbatim}
\usepackage{psfrag}
\usepackage{textcomp}
\usepackage{graphicx}
\usepackage{algorithmic}
\usepackage[Pseudocode]{algorithm}
\usepackage{todonotes}
\usepackage{bm}
\usepackage{empheq}
\usepackage{pgfplots}
\usepackage{multirow}

\pgfplotsset{compat=1.13}

\IEEEoverridecommandlockouts                              

\title{\LARGE \bf
	Insights into the robustness of control point configurations for homography and planar pose estimation
}

\author{Raul Acuna$^{1}$ and Volker Willert$^{1}$
	\thanks{*This work was sponsored by the German Academic Exchange Service (DAAD) and the Becas Chile doctoral scholarship.}
	\thanks{$^{1}$These authors are within the Institute of Automatic Control and Mechatronics, Technische Universit{\"a}t Darmstadt, Germany.{\tt\small (racuna, vwillert)}{\tt\small @rmr.tu-darmstadt.de}}}

\begin{document}
	\maketitle
	\thispagestyle{empty}
	\pagestyle{empty}
	\begin{abstract}
		In this paper, we investigate the influence of the spatial configuration of a number of $n \geq 4$ control points on the accuracy and robustness of space resection methods, e.g. used by a fiducial marker for pose estimation. We find robust configurations of control points by minimizing the first order perturbed solution of the DLT algorithm which is equivalent to minimizing the condition number of the data matrix. An empirical statistical evaluation is presented verifying that these optimized control point configurations not only increase the performance of the DLT homography estimation but also improve the performance of planar pose estimation methods like IPPE and EPnP, including the iterative minimization of the reprojection error which is the most accurate algorithm. We provide the characteristics of stable control point configurations for real-world noisy camera data that are practically independent on the camera pose and form certain symmetric patterns dependent on the number of points. Finally, we present a comparison of optimized configuration versus the number of control points. 
	\end{abstract}
	
	\section{INTRODUCTION}
	The Perspective-n-Point (PnP) problem and the special case of planar pose estimation via homography estimation are some of the most researched topics in the fields of computer vision and photogrammetry. Even though the research in these areas has been wide, there is a surprising lack of information regarding the effect of 3D control point configurations on the accuracy and robustness of the estimation methods.
	
	As shown in Sec.~\ref{state_of_the_art}, it is clear from the literature, that control point configurations are relevant and they do influence the accuracy and robustness of pose estimates. However, the findings are rather general, since they are based on hands-on experience and thus far only lead to some rules of thumb. Most obvious and widely accepted is, that increasing the number of control points increases the accuracy of the results in presence of noise. Further on, in several studies when simulations are performed to compare methods, great care is given to possible singular point configurations, such as non-centered data or near-planar cases which are singularities or degenerate cases for certain estimation methods, so there is a need at least to find out which point configurations are better than others so fair comparisons can be made.
	
	A more thorough evaluation is given for the \textit{normalized} DLT algorithm, whereas the normalization has already shown to improve the estimation because it is related to the condition number of the set of DLT equations \cite{Hartley1997}. The only error analysis for homography estimation found so far by the authors in the literature presents a statistical analysis and simulations of the errors in the homography coefficients \cite{Chen2009}.
	
	\begin{figure}[t]
		\begin{center}
			\includegraphics[width=0.8\columnwidth]{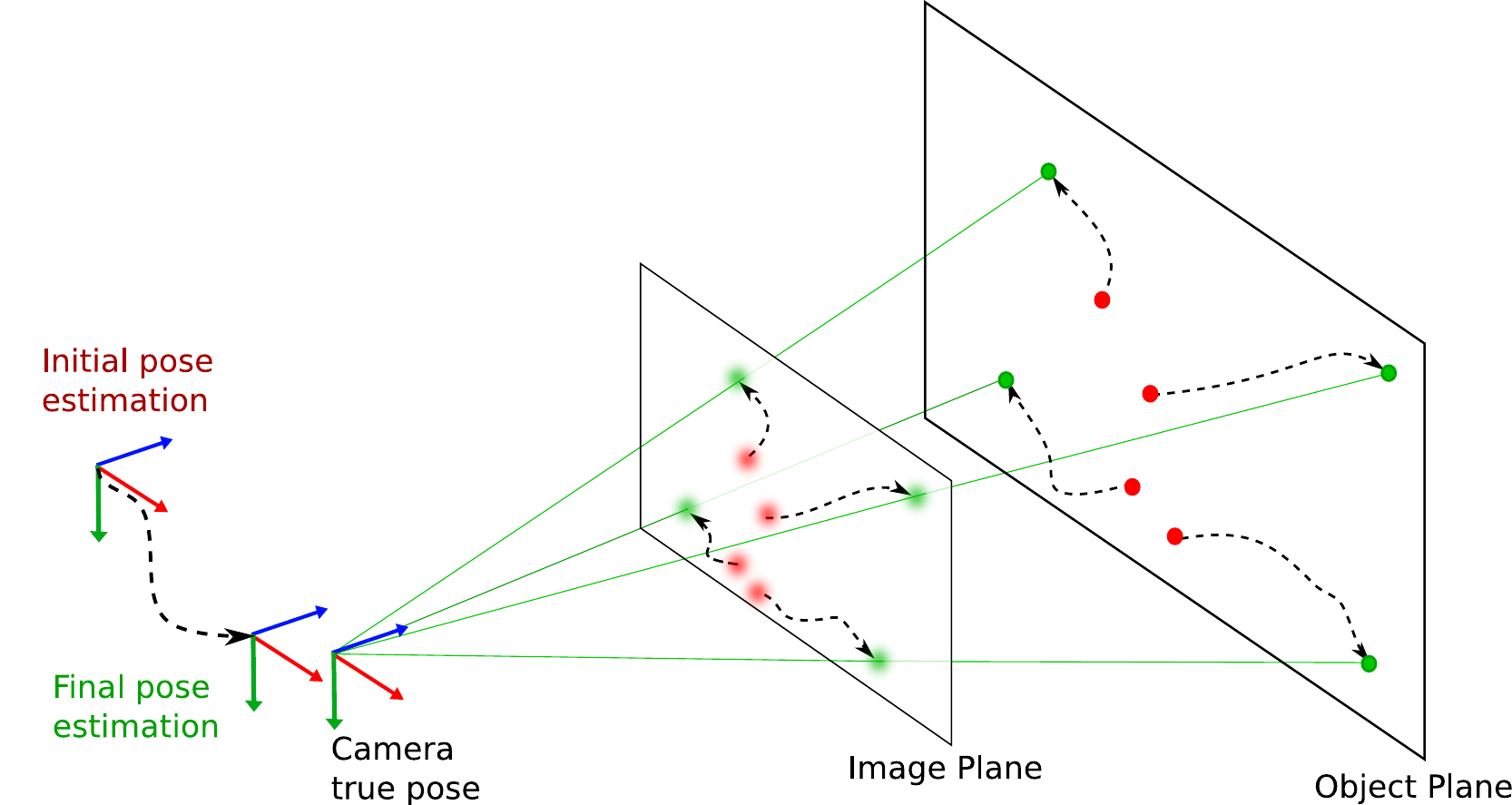}
			\caption{\label{fig:intro_figure}\small Optimizing point configurations: Control points with known 3D coordinates on the object plane (marker) in arbitrary configuration (red) are moved towards optimized configurations (green) for pose estimation from these control points and their corresponding noisy projections on the image plane (blurry red and green). The control points' dynamics \eqref{Eq12} is given by the gradient descent steps minimizing the optimization objective \eqref{Eq11} that results in improved pose estimations (from red to green) close to the true camera pose (black).}
		\end{center}
		\vspace{-0.7cm}
	\end{figure}
	
	However, none of the above give an answer to the question: 
	\textit{Are there optimized perspective-n-point configurations, which can increase the accuracy and robustness of space resection methods?}
	
	If there are, this question includes several follow-up questions: 
	Are the optimized configurations dependent on the pose, or is there only one configuration that is optimal for all poses? What are the specifics of this/these configuration(s) in relation to absolute coordinates and relative distances between coordinates? Are there similarities between configurations that differ in the number of control points? When does an increase in the number of points that are arbitrarily configured outperform the optimal configuration of a small number point set?
	
	In this paper, we search for an answer to these questions in the planar case by proposing an optimization objective to find optimized planar control point configurations. Figure~\ref{fig:intro_figure} sketches the main idea of optimizing the proposed objective via a gradient descent approach and the stepwise improvement of the accuracy of the pose estimate starting from some initial control point configuration. Each descent step leads to a change in control point configuration and thus defines a stable dynamics for the control points that are placed on a planar visual fiducial marker (object plane) converging to stable control point configurations.
	
	The paper is structured as follows:  In Sec.~\ref{state_of_the_art}, we classify pose estimation methods and summarize known findings of control point configurations. In Sec.~\ref{SecBasics}, we derive the optimization objective based on golden standard algorithms for pose estimation. In Sec.~\ref{SimSetup}, we describe the simulation results, and finally, in Sec.~\ref{Conc}, we discuss the results and give conclusions.
	
	\section{State of the art: brief history of space resection and PnP methods}
	\label{state_of_the_art}
	Camera or space resection is a term used in the field of photogrammetry in which the spatial position and orientation of a photo are obtained by using image measurements of control points present on the photo, also know in the computer vision community as the Perspective-n-Point (PnP) problem. PnP can be considered an over-constrained (only for $n \geq 3$) and generic solution to the pose estimation problem from point correspondences. PnP methods can be classified into those which solve for a small and predefined number $n$ of points, and those which can handle the general case. Several solutions have been presented in the literature~\cite{Marchand2016}, which in general provide four solutions for non-collinear points. Thus, prior knowledge has to be included to choose the correct solution. 
	
	Since it has been proven that pose accuracy usually increases with the number of points \cite{Marchand2016}, other PnP approaches that use more points ($n > 3$) are usually preferred. The general PnP methods can be broadly divided into whether they are iterative or non-iterative. Iterative approaches formulate the problem as a non-linear least-squares problem. They differ in the choice of the cost function to minimize, which is usually associated to an algebraic or geometric error. Some of the most important iterative methods in chronological order are: the \textbf{POSIT} algorithm~\cite{Oberkampf1996}, the \textbf{LHM}~\cite{Lu2000}, the Procrustes PnP method or \textbf{PPnP}~\cite{Garro2012} and the global optimization method \textbf{SDP}~\cite{Schweighofer2008}.
	
	Most iterative methods have the disadvantage that they return only a single pose solution, which might not be the true one. Most of them can only guarantee a local minimum and the ones that find a global minimum remain computational intensive. The major limitation of iterative methods is that they are rather slow, neither convergence nor optimality can be guaranteed and a good initial guess is usually needed to converge to the right solution.
	
	Non-iterative methods try to reformulate the problem so it may be solved by a potentially large equation system. However, early non-iterative solvers were also computational demanding and worse for a larger number of points.
	The first efficient and non-iterative $O(n)$ solution was \textbf{EPnP}~\cite{Lepetit2008}, which was later improved by using an iterative method to increase accuracy.
	
	More recent approaches are based on polynomial solvers trying to achieve linear performance without the problems of EPnP and with higher accuracy~\cite{Hesch2011, Li2012,Zheng2013, Kneip2014,Nakano2015}.
	
	A special case of PnP is planar pose estimation, or PPE, which is a space resection problem that involves the process of recovering the relative pose of a plane with respect to a camera's coordinate frame from a single image measurement and which is the focus of this work. A PPE problem can be solved by calculating the object-plane to image-plane homography transformation and then extracting the pose from the homography matrix. This is known as homography decomposition~\cite{Sturm2000, Zhang2000}, or by using a set of points in the plane as the measurement with a special case of the PnP methods (planar PnP). Some of the most important planar PnP methods are the iterative \textbf{RPP-SP}~\cite{Schweighofer2006} and the more recent direct method \textbf{IPPE}~\cite{Collins2014}. In general, planar PnP methods outperform the best homography decomposition methods when noise is present. Additionally, homography decomposition methods only provide a single solution in contrast to modern planar-PnP methods.
	
	The standard linear algorithm for homography estimation is the Direct Linear Transform (DLT)~\cite{RichardHartley2003}, which was improved later in~\cite{Harker2005} using an orthogonalization step. For both methods, the normalization of the measurements is a key step to improve the quality of the estimated homography~\cite{RichardHartley2003}. However, the normalization has some disadvantages~\cite{Rangarajan2009}: First, the normalization matrices are calculated from noisy measurements and are sensitive to outliers, and second, for a given measurement the noise affecting each point is independent of the others. 
	
	
	\subsection{Control points configurations}

	It has been pointed out in the literature ~\cite{Lepetit2008,Li2012} that 3D point configurations have an influence on the local minima of the PnP problem. In the RPnP method paper~\cite{Li2012}, a broad classification of the control points configurations into three groups is presented. The classification is based on the rank of the matrix $\mathbf{M}^T\mathbf{M} \in \mathbb{R}^{3\times 3}$, where $\mathbf{M} = [\mathbf{X}_1, \mathbf{X}_2,\dots,\mathbf{X}_n]^T$, $\mathbf{X}_i$ is the 3D coordinate of control point $i$ and $n$ is the amount of control points.
	
	In EPnP~\cite{Lepetit2008} it is shown that if the control points are taken from the \textit{uncentered data} or the region where the image projections of the control points cover only a small part of the image, the stability of the compared methods greatly degrades. In RPnP it is elaborated that based on the previous classification this \textit{uncentered data} is a configuration that lays between the \textit{ordinary 3D case} and the \textit{planar case}.
	
	Some assumptions about the influence of the control points configurations are also present in the IPPE paper~\cite{Collins2014}. Through statistical evaluations, the authors found out that the accuracy for the 4-point case decreases if the points are uniformly sampled from a given region. They circumvent this problem by selecting the corners of the region as the positions for the control points and then refer the reader to the Chen and Suter paper~\cite{Chen2009}, where the analysis of the stability of the homography estimation to 1st order perturbations is presented. In this analysis, it is clear that the error in homography estimate is dependent on the singular values of the $\mathbf{A}$ matrix in the DLT algorithm (see also next section).
	
	Additionally, in~\cite{Willert2010, Chung2014} 
	evaluations are presented characterizing pose-dependent offsets and uncertainty on the camera pose estimations. It is empirically proven by simulations that some poses of the camera are more stable for the estimation process than others.

	\section{Basics of golden standard algorithms for pose estimation}
	\label{SecBasics}
	
	Before we explain the optimization method for optimizing point configurations, we shortly summarize the \textit{golden standard} optimization methods for
	pose estimation from general and planar point configurations which are the minimization of the reprojection (geometric) error (MRE) for iterative methods and the minimization of the algebraic error for non-iterative methods via the DLT algorithm, respectively. 
	
	\subsection{General point configuration for pose estimation}
	
	Given a 3D-2D point correspondence of $i$-th 3D control point $p_i$ with world $W$ coordinates 
	$\mathbf{X}_i^{W} = [X_i^{W}, Y_i^{W}, Z_i^{W}]^T \in \mathbb{R}^3$ and its corresponding projection onto a planar calibrated camera\footnote{Assuming the calibration matrix $\mathbf{K} \in \mathbb{R}^{3\times 3}$ to be known, the homogeneous image coordinates in pixel 
		$\overline{\mathbf{x}}'_i = [x'_i, y'_i, 1]^T$ can be transformed to homogeneous normalized image coordinates in metric units 
		$\overline{\mathbf{x}}_i = \mathbf{K}^{-1}\overline{\mathbf{x}}'_i$.} with normalized image coordinates $\mathbf{x}_i = [x_i, y_i]^T \in \mathbb{R}^2$ the relation between these points is given by the relative pose\footnote{The rotation matrix is given by: $\mathbf{R} = [\mathbf{r}_1, \mathbf{r}_2, \mathbf{r}_3] \in \mathbb{R}^{3 \times 3} |\,  \mathbf{R}^T\mathbf{R} = \mathbf{I},\, |\mathbf{R}|=1.$ } 
	$g = (\mathbf{R}, \mathbf{T})$ (Euclidean transformation) between world $W$ and camera $C$ frame $\mathbf{X}_i^{C} = \mathbf{R}\mathbf{X}_i^{W}+\mathbf{T}$
	followed by a projection $\pi$ with $\mathbf{x}_i = \pi(\mathbf{X}_i^{C}) = [X_i^{C}/Z_i^{C}, Y_i^{C}/Z_i^{C}]^T$.
	
	This leads to the relation:
	\begin{equation}
	\label{Eq1}
	\mathbf{x}_i = \pi(\mathbf{X}_i^{C}) = \pi(\mathbf{R}\mathbf{X}_i^{W}+\mathbf{T})\,.
	\end{equation}
	
	Including additive noise $\bm{\epsilon}_i = [\epsilon_i, \zeta_i]^T$ on the error-free image coordinates $\mathbf{x}_i$ we get noisy measurements of the image coordinates
	$\tilde{\mathbf{x}}_i = \mathbf{x}_i + \bm{\epsilon}_i$.
	Thus, we can solve for the reprojection error $|\!|\bm{\epsilon}_i|\!|_2^2 = |\!|\tilde{\mathbf{x}}_i - \mathbf{x}_i|\!|_2^2$ of each point
	which is a squared 2-norm. Minimizing the squared 2-norm of all points for the optimal pose $(\hat{\mathbf{R}}, \hat{\mathbf{T}})$ leads to the following least-squares estimator
	\begin{equation}
	\label{Eq2}
	(\hat{\mathbf{R}}, \hat{\mathbf{T}}) = \text{argmin}_{\mathbf{R}, \mathbf{T}} 
	\sum\limits_{i=1}^n |\!|\bm{\epsilon}_i|\!|_2^2\, , \quad n \geq 3 \,.
	\end{equation}
	Iterative gradient descent optimization of \eqref{Eq2} leads to the most accurate pose estimation results in the literature so far,
	also for planar point configurations.
	
	\subsection{Planar points configuration for pose estimation}
	
	If the control points $\mathbf{X}_i^{W}$ are all on a plane $P$, we can define a 
	2D subspace in the 3D world with 
	coordinates\footnote{Corresponding homogeneous coordinates are 
		$\overline{\mathbf{X}}_i^{P} = [X_i^P, Y_i^P, 1]^T \in \mathbb{R}^3$.} 
	$\mathbf{X}_i^{P} = [X_i^P, Y_i^P]^T \in \mathbb{R}^2$.
	Plugging the planar constraint in equation \eqref{Eq1}, extending to homogeneous coordinates and rearranging the equation, leads to an homography mapping
	\begin{equation}
	\mathbf{X}_i^{C} = Z_i^C\overline{\mathbf{x}}_i = \left[\mathbf{r}_1, \mathbf{r}_2, \mathbf{T} \right]\overline{\mathbf{X}}_i^{P}
	= \mathbf{H}\overline{\mathbf{X}}_i^{P}\,.
	\end{equation}
	Eliminating $Z_i^C$, we get $\overline{\mathbf{x}}_i \times \mathbf{H}\overline{\mathbf{X}}_i^{P} = 0$, where
	each point correspondence $\{\mathbf{x}_i, \mathbf{X}_i^P\}$ produces two linearly independent equations
	\begin{equation}
	\mathbf{A}_i\mathbf{h} = 
	\begin{bmatrix}
	\mathbf{O}_{1 \times 3} & -(\overline{\mathbf{X}}_i^P)^T & y_i(\overline{\mathbf{X}}_i^P)^T \\
	(\overline{\mathbf{X}}_i^P)^T & \mathbf{O}_{1 \times 3} & -x_i(\overline{\mathbf{X}}_i^P)^T 
	\end{bmatrix}
	\begin{bmatrix}
	\mathbf{r}_1 \\
	\mathbf{r}_2 \\
	\mathbf{T}
	\end{bmatrix}
	=\mathbf{0}\, ,
	\end{equation}
	with $\mathbf{h}=[\mathbf{r}_1^T, \mathbf{r}_2^T, \mathbf{T}^T]^T \in \mathbb{R}^{9}$ and $\mathbf{A}_i \in \mathbb{R}^{2 \times 9}$.
	
	Again, assuming noisy measurements of the image coordinates
	$\tilde{\mathbf{x}}_i = \mathbf{x}_i + \bm{\epsilon}_i$, we get noisy matrices
	\begin{align}
	\tilde{\mathbf{A}}_i & = 
	\begin{bmatrix}
	\mathbf{O}_{1 \times 3} & -(\overline{\mathbf{X}}_i^P)^T & \tilde{y}_i(\overline{\mathbf{X}}_i^P)^T \\
	(\overline{\mathbf{X}}_i^P)^T & \mathbf{O}_{1 \times 3} & -\tilde{x}_i(\overline{\mathbf{X}}_i^P)^T 
	\end{bmatrix} = 
	\mathbf{A}_i + \mathbf{E}_i \\
	& =
	\begin{bmatrix}
	\mathbf{O}_{1 \times 3} & -(\overline{\mathbf{X}}_i^P)^T & y_i(\overline{\mathbf{X}}_i^P)^T \\
	(\overline{\mathbf{X}}_i^P)^T & \mathbf{O}_{1 \times 3} & -x_i(\overline{\mathbf{X}}_i^P)^T 
	\end{bmatrix} + 
	\begin{bmatrix}
	\mathbf{O}_{1 \times 6} & \zeta_i(\overline{\mathbf{X}}_i^P)^T \\
	\mathbf{O}_{1 \times 6} & \epsilon_i(\overline{\mathbf{X}}_i^P)^T 
	\end{bmatrix}\,. \nonumber
	\end{align}
	
	From $\tilde{\mathbf{A}}_i\mathbf{h} = (\mathbf{A}_i + \mathbf{E}_i)\mathbf{h}$
	we can solve for the algebraic error
	$|\!|\mathbf{E}_i\mathbf{h}|\!|_2^2 = |\!|(\tilde{\mathbf{A}}_i-\mathbf{A}_i)\mathbf{h} |\!|_2^2= |\!|\tilde{\mathbf{A}}_i\mathbf{h}|\!|_2^2$ of each point, because $\mathbf{A}_i\mathbf{h} = \mathbf{0}$ holds.
	Minimizing the squared 2-norm of all points for the optimal homography $\hat{\mathbf{h}}$ 
	leads to the following least-squares estimator
	\begin{equation}
	\label{Eq5}
	\hat{\mathbf{h}} = \text{argmin}_{\mathbf{h}} 
	\sum\limits_{i=1}^n |\!|\mathbf{E}_i\mathbf{h}|\!|_2^2\, , \quad s.t. \,\,\, |\!|\mathbf{h}|\!|=1\,,\quad n \geq 4 \,.
	\end{equation} 
	Since $\mathbf{h}$ contains 9 entries, but is defined only up to scale the total number of degrees of freedom is 8. Thus, the additional constraint $|\!|\mathbf{h}|\!|=1$ is included to solve the optimization.
	
	Now, stacking all $\{\tilde{\mathbf{A}}_i\}$ and $\{\mathbf{E}_i\}$ as $\tilde{\mathbf{A}}=[\tilde{\mathbf{A}}_1^T, \dots, \tilde{\mathbf{A}}_n^T]^T \in \mathbb{R}^{2n \times 9}$
	and $\mathbf{E}=[\mathbf{E}_1^T, \dots, \mathbf{E}_n^T]^T \in \mathbb{R}^{2n \times 9}$ respectively, we arrive at solving
	the noisy homogeneous linear equation system
	\begin{equation}
	\label{Eq6}
	\tilde{\mathbf{A}}\mathbf{h}=\mathbf{E}\mathbf{h} \,.
	\end{equation}
	
	The solution of \eqref{Eq6} is equivalent to the solution of \eqref{Eq5} 
	and is given by the DLT algorithm applying a singular value decomposition (SVD) of 
	$\tilde{\mathbf{A}} = \tilde{\mathbf{U}}\tilde{\mathbf{S}}\tilde{\mathbf{V}}^T$,
	whereas $\hat{\mathbf{h}}=\tilde{\mathbf{v}}_9$ with $\tilde{\mathbf{v}}_9$ being the 
	right singular vector of $\tilde{\mathbf{A}}$, associated with the least singular value $\tilde{s}_9$. Usually, an additional normalization step of the coordinates of the control points and its projections is performed leading to the normalized DLT algorithm which is the golden standard for non-iterative pose estimation, because it is very easy to handle and serves as a basis for other non-iterative as well as iterative pose estimation methods.
	
	\section{Optimizing points configuration for pose estimation}
	\label{IdeaPointConfigSearch}
	
	In order to find an optimal control points configurations, we need a proper optimization criterion. 
	In the following, we propose an optimization criterion that is optimal for planar pose estimation using the (normalized) DLT algorithm, since it is the critical first step in planar pose estimation methods (even the gold standard of the minimization of the reprojection error requires a good initial guess, which is obtained from the DLT).
	We start with availing ourselves of perturbation theory applied to singular value decomposition of noisy matrices \cite{Stewart1998} and have a look at the first order perturbation expansion for the perturbed solution of the DLT algorithm, given in \cite{Chen2009}, which is 
	\begin{equation}
	\label{Eq10}
	\hat{\mathbf{h}}=\tilde{\mathbf{v}}_9 = \mathbf{v}_9 - \sum\limits_{k=1}^8 \frac{\mathbf{u}_k^T\mathbf{E}\mathbf{v}_9}{s_k} \mathbf{v}_k\,.
	\end{equation}
	Equation \eqref{Eq10} clearly shows that the optimal solution for the homography that equals the 
	right singular vector of the unperturbed matrix $\mathbf{A}$, associated with the least singular value\footnote{The singular values are arranged in descending order: $s_1 \geq s_2 \geq \dots \geq s_8 \geq s_9 = 0$.}
	$s_9 = 0$, is perturbed by the second term in \eqref{Eq10}. The second term is a weighted sum of the first eight optimal right singular vectors $\mathbf{v}_k$, whereas the weights $\mathbf{u}_k^T\mathbf{E}\mathbf{v}_9/ s_k$ are the influence of the measurement errors $\mathbf{E}$
	on the unperturbed solution $\mathbf{v}_9$ along the different $k$ dimensions of the model space.
	The presence of very small $s_k$ in the denominator can give us very large weights for the corresponding model space basis vector $\mathbf{v}_k$ and dominate the error. Hence, small singular values $s_k$ cause the estimation $\hat{\mathbf{h}}$ to be extremely sensitive to small amounts of noise in the data and correlates with the singular value spectrum\footnote{Here, the singular value spectrum between the first and second-last singular value is relevant, because $s_9=0$ holds.} $(s_1-s_8)$ as follows: The smaller the singular value spectrum, the less perturbed the estimation is. It is also well known, that the condition number of a matrix with respect to the 2-norm is given by the ratio between the largest and, in our case, second-smallest singular value \cite{Golub2013}
	\begin{equation}
	\label{Eqc}
	c(\mathbf{A}) = \| \mathbf{A} \|_2\| \mathbf{A}^{-1} \|_2 = \frac{s_{max}}{s_{min}}= \frac{s_1}{s_8}\,,
	\end{equation}    
	which is minimal if the singular value spectrum is minimal.
	The normalization of the control points and its projections which leads to the normalized DLT algorithm
	has already shown to improve the condition of matrix $\mathbf{A}$ \cite{Hartley1997}.
	Thus, we simply try to minimize the condition number $c$ of matrix $\mathbf{A}$
	with respect to all $n$ control points $\{\mathbf{X}_i^P\}$ like follows:
	\begin{equation}
	\label{Eq11}
	\{\hat{\mathbf{X}}_i^P\} = \text{argmin}_{\{\mathbf{X}_i^P\}} c\left(\mathbf{A}(\{\mathbf{X}_i^P\})\right) \,.
	\end{equation}
	Optimization of \eqref{Eq11} is realized with a gradient descent minimization,
	whereas for calculation of the gradient vector we use
	automatic differentiation\footnote{For implementation, we used \textit{autograd} \cite{Maclaurin2016}.} \cite{Rall1981}.
	This leads to the final discrete control points dynamics
	\begin{equation}
	\label{Eq12}
	\mathbf{X}_i^P(t+1) = \mathbf{X}_i^P(t) - \alpha(t) \nabla c\left(\mathbf{A}\left(\mathbf{X}_i^P(t)\right)\right)\,,
	\end{equation}
	for each iteration $t$ and stepsize $\alpha(t)$, which is adapted using SuperSAB \cite{Tollenaere1990}.
	The control points dynamics can now be used to find optimal control point configurations for pose estimation from planar markers.

	
	Given perturbations on the matrix $\mathbf{A}$ the relative error on the estimation of the homography parameters is defined as $\mathbf{\xi} = (\mathbf{h} - \hat{\mathbf{h}})/\hat{\mathbf{h}}$ and from perturbation theory the following inequality defines an upper bound for the relative error:
	
	\begin{equation}
	\label{Eq_inequality}
	|\!|\mathbf{\xi}|\!| \leq c(\mathbf{A})|\!|\mathbf{A}-\tilde{\mathbf{A}}|\!| / |\!|{\mathbf{A}}|\!|\,. 
	\end{equation}
	
	To find a lower bound, we can use the error of using a perturbed matrix $\tilde{\mathbf{A}}$ with the true homography $\mathbf{h}$ defined as $\tilde{\mathbf{A}}\mathbf{h}$ and the error of using the optimal homography estimation $\hat{\mathbf{h}}$ with the same perturbed matrix defined as $\tilde{\mathbf{A}}\hat{\mathbf{h}}$ to build the following inequality:
	
	\begin{equation}
	|\!|\tilde{\mathbf{A}}\mathbf{h} - \tilde{\mathbf{A}}\hat{\mathbf{h}}|\!|	= 	|\!|\tilde{\mathbf{A}}(\mathbf{h}-\hat{\mathbf{h}})|\!| \leq |\!|\tilde{\mathbf{A}}|\!||\!|\mathbf{h}-\hat{\mathbf{h}}|\!|,
	\end{equation}
	
	which then divided by $|\!|\hat{\mathbf{h}}|\!|$ leads to a lower bound of the relative error:
	\begin{equation}
	|\!|\mathbf{\xi}|\!| \geq |\!|\tilde{\mathbf{A}}(\mathbf{h} - \hat{\mathbf{h}})|\!|  / (|\!|\tilde{\mathbf{A}}|\!||\!|\hat{\mathbf{h}}|\!|)\,. 	
	\end{equation}

	The upper bound implies that there are only two ways to improve the maximum values of the relative error, either by reducing the perturbation of the $\mathbf{A}$ matrix, which usually can't be controlled, or by improving the condition number of the $\mathbf{A}$ matrix, which can be done by a normalization step in the DLT transform and by selecting optimized control point configurations. 
	
	\section{Simulation and real experiment results}
	\label{SimSetup}
	Our simulation setup is based on a perspective camera model and a planar visual marker on $Z^W_i=0$ centered in the origin $\mathbf{X}^W_o =[0,0,0]^T$ of world coordinates, we impose an arbitrary circular limit with a radius of $r=0.15$ meters, this allows a smooth movement of the control points during the optimization while keeping them inside camera image. Rectangular limits were also tested but the discontinuities on the corners restrict the movement of the points.
	
\begin{figure}[t]
		\begin{center}
			\includegraphics[width=0.8\columnwidth]{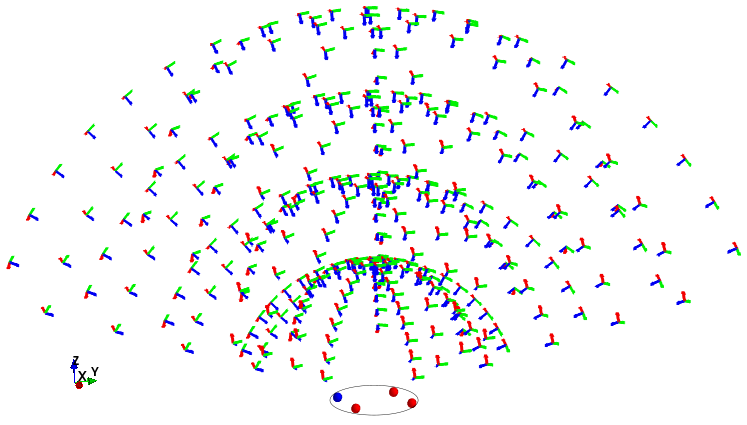}
			\caption{\label{fig:camera_poses} \small Distribution of 400 camera poses used in simulations. A limiting circular plane including $n$ control points is displayed at the bottom. The cameras are distributed evenly on spheres of evenly sampled radii, each one looking at the center of the circular plane keeping the complete circular plane in field of view.}
		\end{center}
		\vspace{-0.7cm}
	\end{figure}

	A set of control points are randomly defined inside the limits of this circular plane, which are then projected onto the camera image\footnote{Camera parameters: size $640 \times 480\,[pixel^2]$, intrinsic parameters $\mathbf{K}=[800, 0, 320 ; 0 , 800 , 240 ; 0 , 0 , 1]$.}. A uniform distribution of 400 camera poses is defined around the marker as displayed in Fig. \ref{fig:camera_poses}, this distribution provides a wide combination of rotation and translations (without lack of generalization) in the whole range of detection and allows us to properly compare the final point configurations in image coordinates.

	\subsection{Evaluations}   
	
	To evaluate the improvement of the gradient descent optimization, we consider the optimization objective, which is the condition number \eqref{Eqc} at each iteration $t$, given by $c(\mathbf{A}(t))$ in the DLT algorithm. To evaluate the effect of the optimization \eqref{Eq11} on the underlying homography estimate $\hat{\mathbf{H}}(t)$ using a given set of $n$ control points $\{\mathbf{X}^P_{i}\}(t)$, the movement of the points during the optimization is constrained to the limits of the circular bounds, we rely on the reprojection error $HE(\hat{\mathbf{H}}(t))$ induced by the estimated homography $\hat{\mathbf{H}}(t)$ given by
	
	\begin{equation}
	HE\left(\hat{\mathbf{H}}(t)\right) = \frac{1}{M}
	\sum\limits_{j=1}^M |\!| \mathbf{x}_j(t) - \pi\left(\hat{\mathbf{H}}(t)\overline{\mathbf{X}}_j^{P}(t)\right)|\!|_2^2\,,
	\end{equation}
	for a fixed set of $M$ validation control points $\{\mathbf{X}_j^{P}\} \not\in \{\mathbf{X}^P_{i}\}(t)$ that are evenly distributed on the object plane covering an area larger than the limits of the circle. Thus, it is possible to measure how good $\hat{\mathbf{H}}(t)$ is able to represent the true homography $\mathbf{H}$ beyond the space of the control points.   
	
	Each simulation for a given camera pose is then performed in the following way: 
	1) An initial random $n$-point set $\{\mathbf{X}^P_{i}\}(t_{start})$ is defined inside the circular plane 
	2) For each iteration step $t$ an improved set of control points $\{\mathbf{X}^P_{i}\}(t)$ is obtained by \eqref{Eq12} and projected to image coordinates $\{\mathbf{x}_{i}\}(t)$ using the true camera pose ${\mathbf{R},\mathbf{T}}$ and calibration matrix $\mathbf{K}$. 
	Then, the correspondences $\{\mathbf{x}_{i}(t), \mathbf{X}^P_{i}(t)\}$ are used to calculate $\mathbf{A}(t)$ and $c(\mathbf{A}(t))$. 
	3) For each $t$ a statistically meaningful measure of the homography estimation robustness against noise is desired. Thus, 1000 runs of the homography estimation using the normalized DLT algorithm were performed\footnote{The homography estimation method presented in \cite{Harker2005} and the gradient based one of OpenCV were also tested. The results almost do not differ for low point configurations to the DLT, so it was the chosen one for the experiments.}. In each of these runs Gaussian noise with standard deviation $\sigma_G$ was added to the image coordinates for the simulation of real camera measurements
	$\{\tilde{\mathbf{x}}_{i}\}(t)$. Finally, the error $HE\left(\hat{\mathbf{H}}(t)\right)$ is calculated in each run and the average $\mu\left(HE\left(\hat{\mathbf{H}}(t)\right)\right)$ and standard deviation $\sigma\left(HE\left(\hat{\mathbf{H}}(t)\right)\right)$ of this error for all runs is computed.
	
\begin{figure}[t]
		\begin{center}
			\includegraphics[width=0.8\columnwidth]{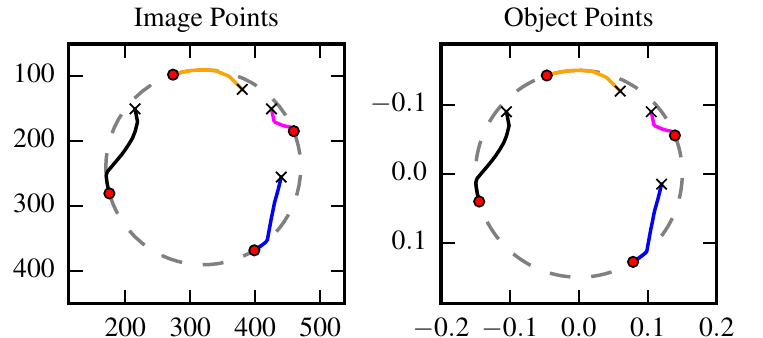}
			\caption{\label{fig:FP_points}\small (Fronto-Parallel). Movement of control points in image and object coordinates during optimization for a fronto-parallel camera configuration simulation until an optimized configuration (red dots) limited by the circle (dashed grey line). }
		\end{center}
		\vspace{-0.5cm}
	\end{figure}

	
	As illustration of the gradient minimization process an example case of a simulation in a fronto-parallel camera pose for a 4-point configuration is presented. A Gaussian noise of $\sigma_G=4$ pixel is added to image coordinates for the homography estimation runs. In Fig. \ref{fig:FP_points} the initial object and image point configurations are shown. 
	
\begin{figure}[t]
		\begin{center}
			\includegraphics[width=0.9\columnwidth]{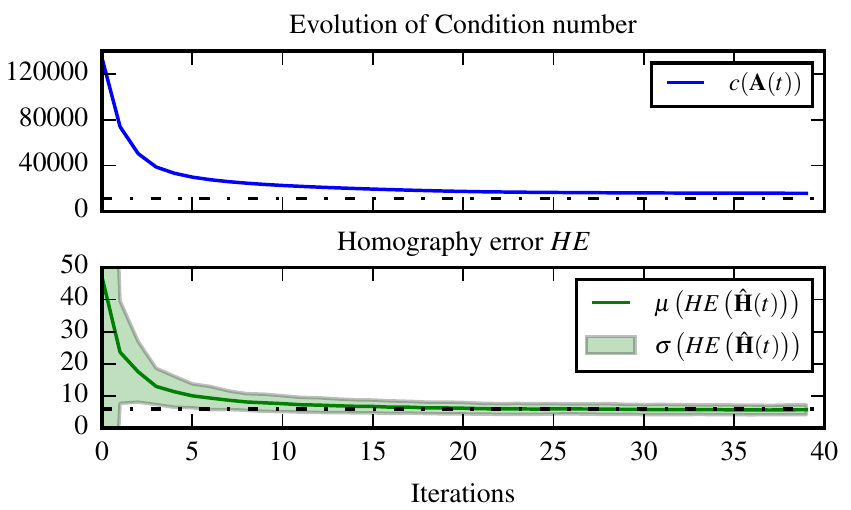}
			\caption{\label{fig:FP_cond_homo_error}\small (Fronto-Parallel). Evolution of the condition number $c(\mathbf{A}(t))$ as well as mean $\mu$ and standard deviation $\sigma$ of the homography reprojection error $HE\left(\hat{\mathbf{H}}(t)\right)$ during gradient descent. For comparison, the dashed-dotted black line represents the mean value for an ideal 4-point square.}
		\end{center}
		\vspace{-0.5cm}
	\end{figure}

	The evolution of $c(\mathbf{A}(t))$ as well as $\mu\left(HE\left(\hat{\mathbf{H}}(t)\right)\right)$ and $\sigma\left(HE\left(\hat{\mathbf{H}}(t)\right)\right)$
	is presented in Fig. \ref{fig:FP_cond_homo_error}. The condition number decreases drastically in the first iterations of the gradient descent, and by doing so the mean and standard deviation of $HE\left(\hat{\mathbf{H}}(t)\right)$ is also reduced. With more iterations both metrics slowly and smoothly converge to a stable minimum value. 
	
	This first result in itself is highly representative as it proves that some point configurations increase the accuracy of homography estimation methods as well as the robustness to noise and it is also possible to obtain optimized point configurations (which are better than random ones).

	
	\begin{figure}[t]
		\begin{center}
			\includegraphics[width=0.9\columnwidth]{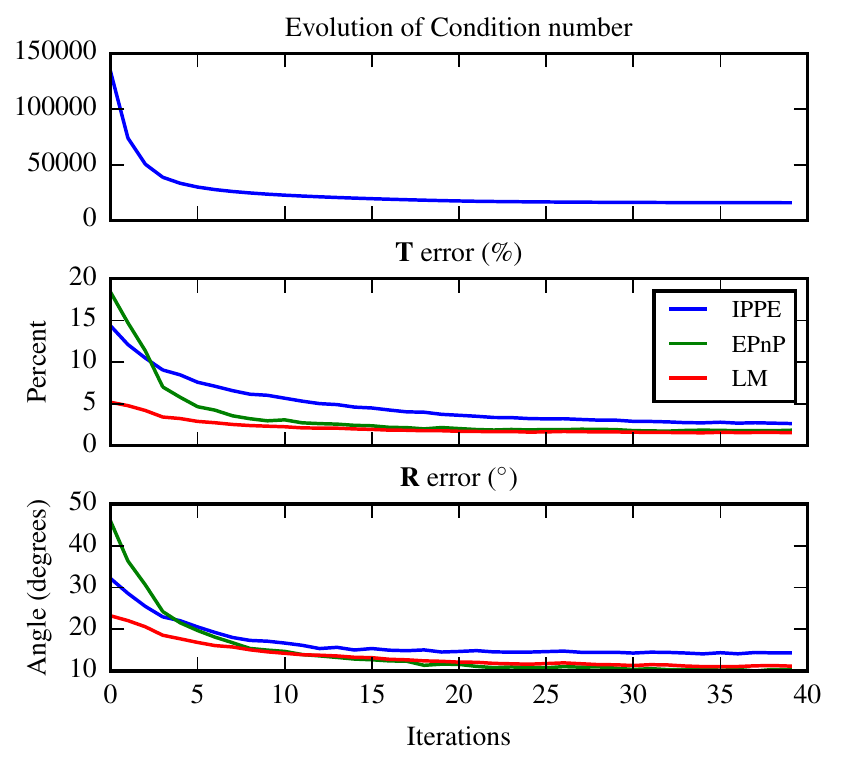}
			\caption{\label{fig:FP_pnp_results_global}\small  (Fronto-Parallel). Comparison of the evolution of the mean errors for different PnP estimation methods during the iterative optimization process. The initial points were the same for all runs.}
		\end{center}
		\vspace{-0.5cm}
	\end{figure}
	
	\begin{figure}[t]
		\begin{center}
			\includegraphics[width=\columnwidth]{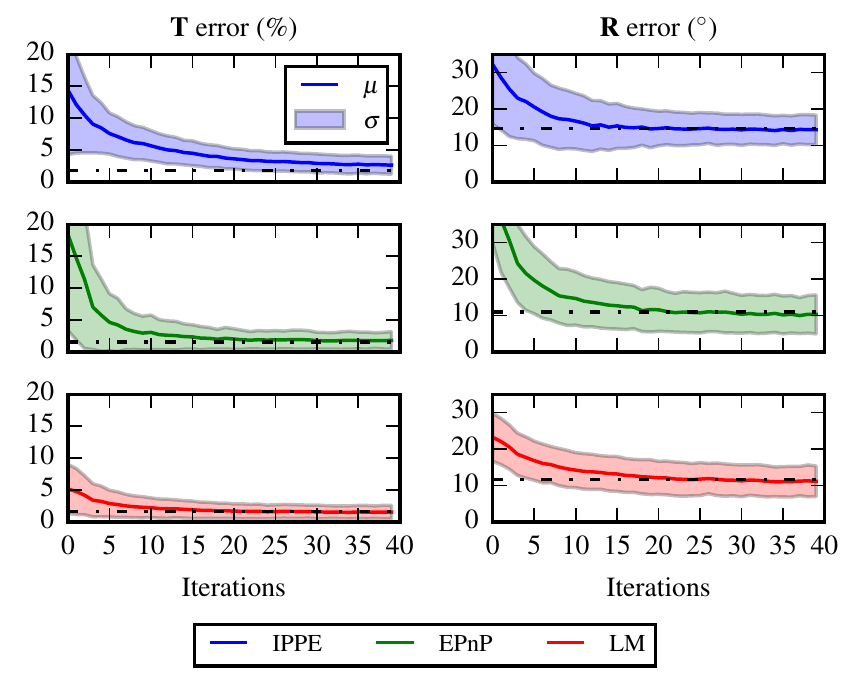}
			\caption{\label{fig:FP_pnp_results_detailed}\small(Fronto-Parallel). Mean values (colored lines) and standard deviations (filled colored areas) of translational $\mathbf{T}$
				and rotational $\mathbf{R}$ error for each method. The dashed-dotted black line represents the mean value for an ideal 4-point square.}
		\end{center}
		\vspace{-0.5cm}
	\end{figure}
	
	%
	%
	%

	\begin{figure}[t]
		\begin{center}
			\includegraphics[width=0.8\columnwidth]{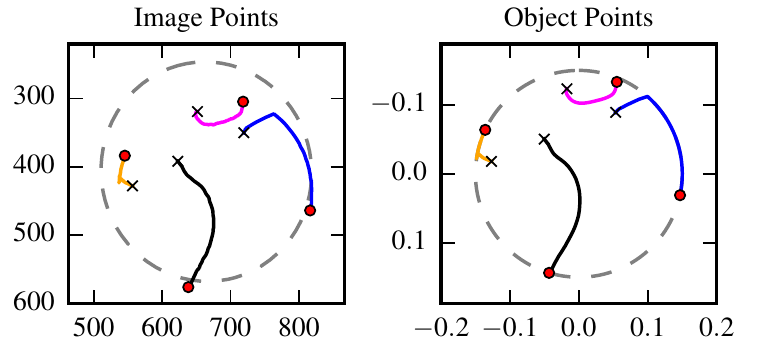}
			\caption{\label{fig:RE_points} \small (Real). Movement of control points in image and object coordinates during gradient descent for the experiment with a real camera. See our video for further details.}
		\end{center}
		\vspace{-0.5cm}
	\end{figure}
	
	\begin{figure}[t]
		\begin{center}
			\includegraphics[width=0.9\columnwidth]{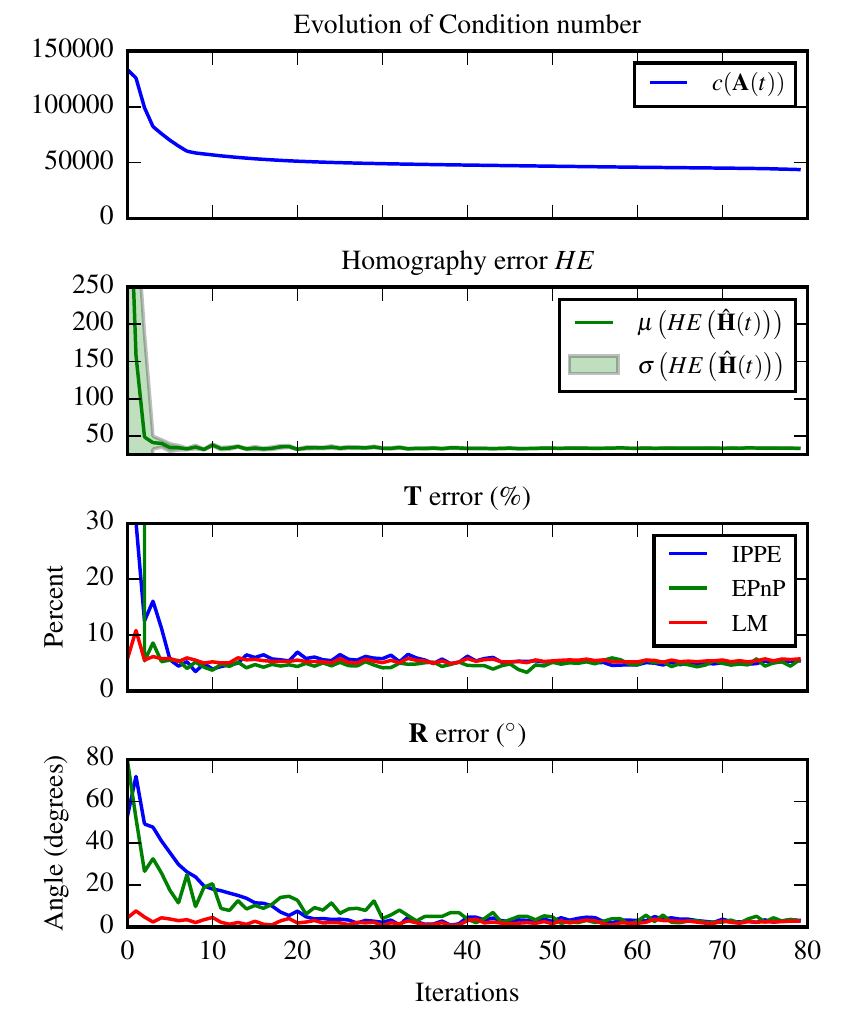}
			\caption{\label{fig:RE_homo_pnp_results_global} \small(Real). Evolution of the condition number and the homography reprojection error during gradient descent using a real camera.}
		\end{center}
		\vspace{-0.75cm}
	\end{figure}

	\begin{figure}[t]
		\begin{center}
			\includegraphics[width=\columnwidth]{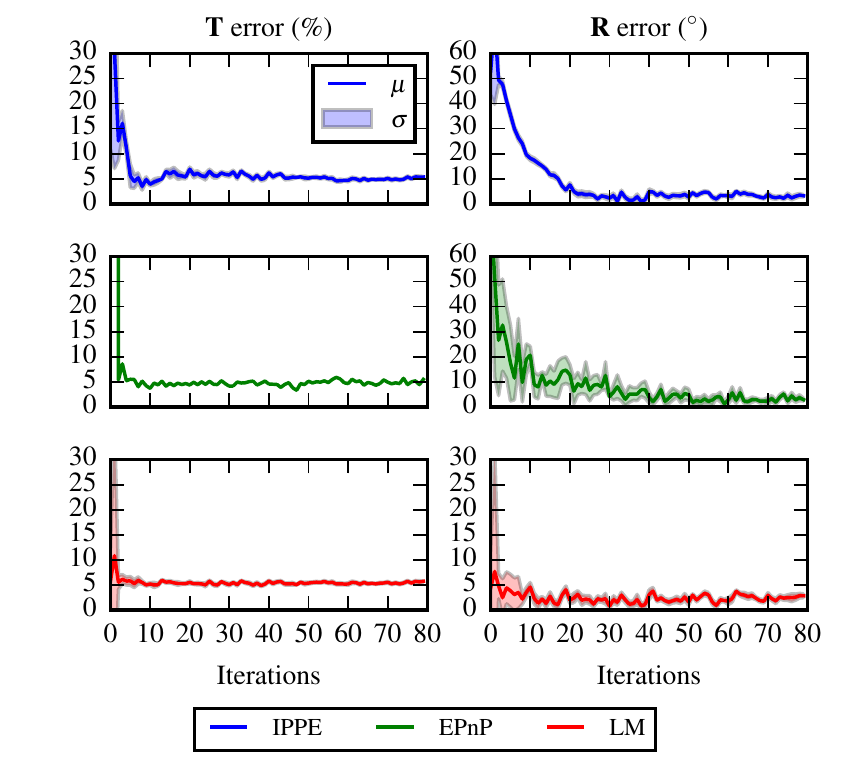}
			\caption{\label{fig:RE_pnp_results_detailed}\small (Real). Detailed view of the standard deviation of each method represented by the filled, lightly colored areas.}
		\end{center}
		\vspace{-0.5cm}
	\end{figure}

	Motivated by the homography results, it was of interest to test if the optimization of control point configurations could improve as well the accuracy of pose estimation algorithms. Thus, three different pose estimation algorithms\footnote{For the EPnP and LM methods, the OpenCV implementations were used, and for IPPE the Python implementation provided in the author's github repository.} were run at each iteration $t$ of the optimization process, namely: 1) a non-iterative PnP method \textbf{EPnP}~\cite{Lepetit2008}, 2) a planar pose estimation method \textbf{IPPE}~\cite{Collins2014}, and 3) an iterative one based on the Levenberg-Marquardt optimization denoted as \textbf{LM}. 
	
	As in similar works \cite{Lepetit2008,Collins2014}, we denote $\left(\hat{\mathbf{R}}(t), \hat{\mathbf{T}}(t)\right)$ as the estimated rotation and translation for a given camera pose at iteration $t$ and by $(\mathbf{R}, \mathbf{T})$ the true rotation and translation. The error metrics for pose estimation are defined as follows:
	\begin{itemize}
		\item  RE$\left(\hat{\mathbf{R}}(t)\right)$ is the rotational error (in degrees) defined as the minimal rotation needed to align $\hat{\mathbf{R}}(t)$ to $\mathbf{R}$. It is obtained from the axis-angle representation of $\hat{\mathbf{R}}(t)^T\mathbf{R}$.
		
		\item TE$\left(\hat{\mathbf{T}}(t)\right) = \|\hat{\mathbf{T}}(t) - \mathbf{T}\|_2/\|\mathbf{T}\|_2\times 100 \%$ is the relative error in translation. 
	\end{itemize}
	
	Similar to the homography simulation, for each iteration $t$, 1000 runs of the pose estimation with noisy correspondences for each of the PnP methods were performed. Then, the mean and standard deviation of RE and TE for the 1000 runs were calculated for each iteration. 
	The PnP simulation results for the fronto-parallel case are presented in Fig.~\ref{fig:FP_pnp_results_global} comparing the performance of all methods together and in Fig.~\ref{fig:FP_pnp_results_detailed} details about the standard deviation of each method are shown. 
	
	
	
	A real experiment was also implemented in order to test if the simulation assumptions (Gaussian image noise and perfect intrinsics) may affect the results in practical applications\footnote{A video of this experiment: https://youtu.be/a6lDrwgqNmY.}. A computer screen was used as the planar fiducial marker to dynamically display the points during gradient descent. A set of 4 circles was displayed for each iteration of the optimization. These circles were then captured by a PointGrey Blackfly camera\footnote{Camera parameters: size $1288 \times 964\,[pixel^2]$, intrinsic parameters $\mathbf{K}=[1070.82, 0, 647.98 ; 0 , 1071.20 , 488.27 ; 0 , 0 , 1]$.} and detected using a Hough transform based circle detector. We performed 100 detections for each gradient descent iteration. An Optitrack system was used to measure the ground truth pose of the camera relative to the marker screen. 
	The results of running the optimization process for a set of 4 random initial points are shown in figures~\ref{fig:RE_points}, \ref{fig:RE_homo_pnp_results_global} and \ref{fig:RE_pnp_results_detailed}. 
	
\begin{figure}[t]
		\begin{center}
			\includegraphics[width=\columnwidth]{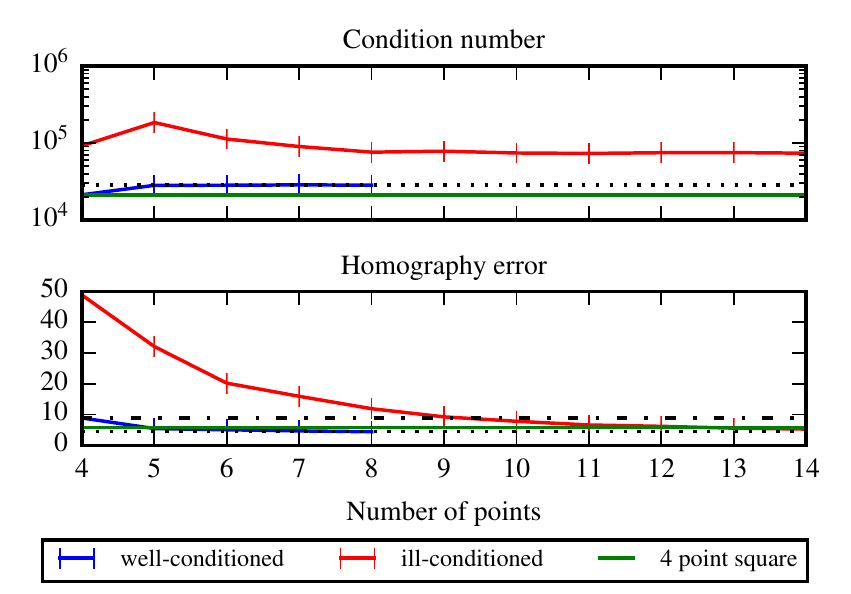}
			\caption{\label{fig:comp_homo}\small Robustness (cond. num.) and accuracy (homography error) dependent on the number of points for well- and ill-conditioned point configurations as well as an ideal 4-point square (green line).}
		\end{center}
		\vspace{-0.5cm}
	\end{figure}

\begin{figure}[t]
		\begin{center}
			\includegraphics[width=\columnwidth]{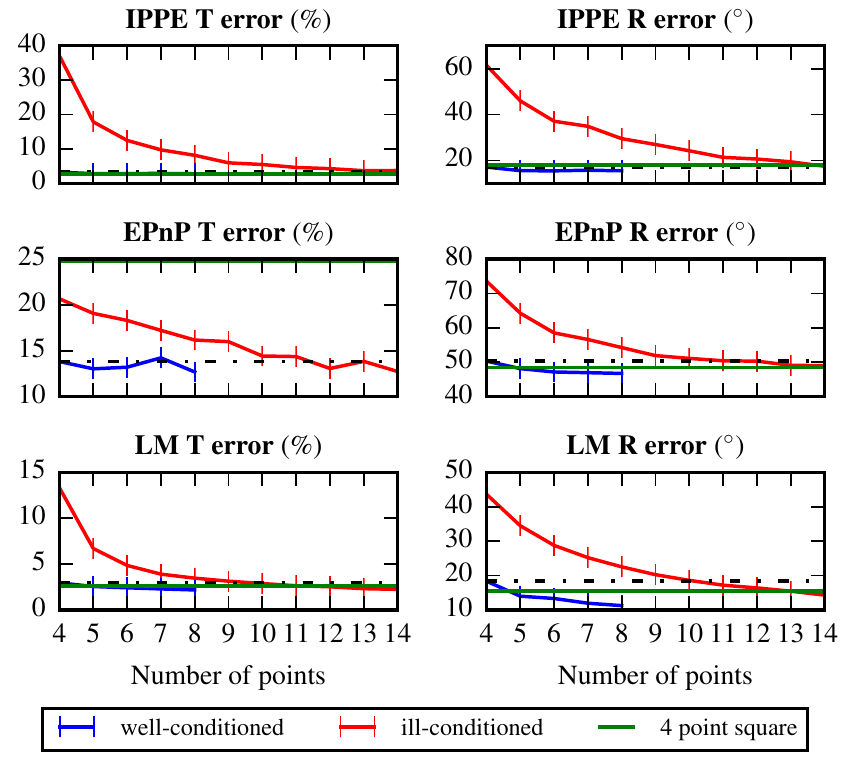}
			\caption{\label{fig:comp_pose}\small Comparison of different pose stimation methods for different numbers of control points for well- and ill-conditioned point configurations as well as an ideal 4-point square (green line).}
		\end{center}
		\vspace{-0.5cm}
	\end{figure}

\begin{figure}[t]
		\begin{center}
			\includegraphics[width=0.7\columnwidth]{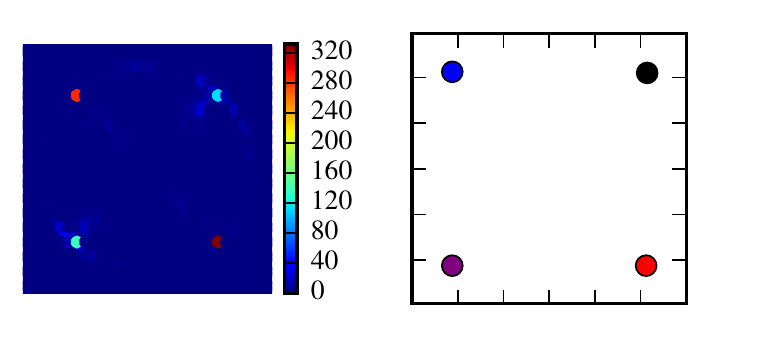}
			\caption{\label{fig:final_points4}\small Left: 2D histogram of final 4-point configurations for all camera poses. Right: One representative final point configuration. In object coordinates.}
		\end{center}
		\vspace{-0.5cm}
	\end{figure}

\begin{figure}[t]
		\begin{center}
			\includegraphics[width=\columnwidth]{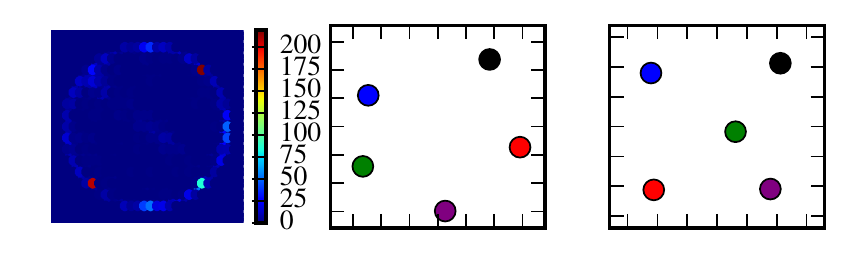}
			\caption{\label{fig:final_points5}\small Left: 2D histogram of final 5-point configurations for all camera poses. Middle/right: Two representative final point configurations. In object coordinates.}
		\end{center}
		\vspace{-0.8cm}
	\end{figure}

	Next, the relationship between badly configured points and optimized points was studied. For each camera pose in the distribution of Fig. \ref{fig:camera_poses}, 100 different initial random $n$-point configurations  with $n \in \{4,5,6,7,8\}$ were simulated and the optimization process was performed. In this case, only the initial and final values of the point configuration metrics are stored. Thus, it is possible to compare the methods based on \textit{ill-conditioned} (random initial points) and \textit{well-conditioned} (after optimization) point configurations. In Fig.~\ref{fig:comp_homo} the results for the homography estimation are presented and in Fig.~\ref{fig:comp_pose} the results of the pose estimation. Finally, in figures \ref{fig:final_points4} and \ref{fig:final_points5}, the final point configurations for all the camera poses are shown as a 2D histogram and some example configurations are shown for the 4-point and 5-point case.

	\subsection{Discussion}

	
	In the results, it is observed that control point configurations have a strong effect on the accuracy of homography and planar PnP methods. There are indeed optimized configurations which are better than random and it is possible to find them using out method.

	For the 4-point case, our empirical results show that a square-like shape is the most common minima and a very stable and robust configuration for all camera poses (see Fig.~\ref{fig:final_points4}) and as shown on Fig.~\ref{fig:final_points5} even for the 5-point case the corners of a square-like shape are common. The optimized point configurations do not show any strong dependency with the pose of the camera (besides scale and image limits), it is mainly related to the distribution of the points in camera image coordinates since they are driven to distribute in space and they tend to increase the distance to each other.

	On the first iterations of the optimization is when the increase in accuracy is stronger, which means that the condition number is a good optimization objective. For example, the improvement in accuracy from a square-like configuration to a perfect square is very small, but the increase of accuracy from random points to the square-like shapes obtained on the first iterations of the optimization is radical. 
	
	The smaller the number of control points the more is the relative improvement on the estimates for all of the evaluated methods. For example, the accuracy using 4 points is always better than random point configurations with more points $4 < n \leq 9$ as can be seen in Fig.~\ref{fig:comp_homo} for homography and Fig.~\ref{fig:comp_pose} for PnP. Thus, the configuration of the control points has more effect on the accuracy than the number of control points.

	The improvement in the EPnP and IPPE methods is more pronounced than for LM, which is in itself an interesting result since those methods take considerable less computation time. For well-configured points, the methods converge to similar error values (see Fig.~5, 6 and Fig.~8) and both mean and variance are reduced, this means that well-conditioned points can be used for fair comparison of pose estimation algorithms. LM also has increased accuracy although our optimization objective is not directly related to the minimization of the reprojection error, this shows the importance of having a good initial guess. The results of the real experiment closely match the simulations.


	
	
	
	\section{Conclusions and future work}
	\label{Conc}
	A method for obtaining optimized control points for homography estimation is presented. The lower the number of control points the more the point configuration has an influence on the accuracy of homography and PnP estimation methods. Our empirical results show that a square is a very stable and robust configuration for all camera poses. Optimized points configurations follow simple rules, they are driven to distribute in space and they tend to increase the distance to each other, this includes the optimized 4 point configuration as a subset. Finally, we found that there is almost no difference in accuracy between IPPE and LM when optimized point configurations are used. In future work, we will try to generalize the results to non-planar point configurations and use other optimization metrics such as the trace of the posterior covariance matrix in the reprojection error which is commonly used in the optimal sensor placement research field.  

	{\small
		\bibliographystyle{ieee}
		\bibliography{IEEEabrv,bib_icra2018,volker}
	}
\end{document}